\pdfoutput=1
\documentclass[runningheads]{llncs}

\usepackage[T1]{fontenc}
\usepackage{lmodern}
\usepackage{graphicx}
\usepackage{booktabs}
\usepackage{tabularx}
\usepackage{array}
\usepackage{amsmath}
\usepackage{amssymb}
\usepackage{url}
\usepackage{multirow}
\usepackage{placeins}

\newcommand{\sparseftwo}{Val.-$F_2$}

\renewcommand{\arraystretch}{0.86}

\begin{document}

\title{Sparse Incident-Cluster Learning for 12-hour Port Flood Pre-warning in Digital-Twin Analytics}

\titlerunning{Sparse Incident-Cluster Learning}

\author{Jie Zhang\inst{1}, Qiang Ni\inst{1}, David Windridge\inst{2} \and Huan X. Nguyen\inst{2}}

\authorrunning{J. Zhang, Q. Ni, D. Windridge, H. Nguyen}

\institute{
Lancaster University, Lancaster, United Kingdom\\
\email{\{j.zhang110, q.ni\}@lancaster.ac.uk}\\
Middlesex University, London, United Kingdom\\
\email{\{d.windridge, h.nguyen\}@mdx.ac.uk}
}

\maketitle

\begin{abstract}
Port flood digital twins require pre-warning analytics that can support decisions before operational disruption, yet the number of independent official warning incidents is often small and adjacent observations are temporally dependent. Under these conditions, row-level classification can overstate performance by distributing related windows from the same warning episode across model-development and evaluation subsets. This paper formulates 12-hour port flood pre-warning as an incident-cluster learning problem in which sparsity is defined at the event level.

We evaluate a digital-twin analytics module that represents recent observations as eight-point local water-level histories, prediction-time contextual covariates, and physically interpretable short-window dynamic descriptors. The protocol combines fold-specific sparse feature selection, warning-cluster grouping, negative-label controls, 100-repeat random top-$k$ controls, and alert-episode evaluation. Liverpool is the primary four-cluster case study; harmonised Humber/Hull-proxy and Wessex South datasets provide protocol-transfer checks.

Across the four Liverpool folds, the fold-specific top-10 ElasticNet model attains a mean $F_2$ of 0.696, compared with 0.633 for an ElasticNet model without top-$k$ truncation and 0.681 for the full-feature weighted XGBoost reference. This is the strongest mean $F_2$ among the ElasticNet variants and is competitive with the nonlinear reference while using ten selected predictors. Complete 100-repeat random controls place the selected result above the repeat-level 95th percentile of broad and same-family random subsets. Contextual covariates provide a strong prediction-time anchor, complemented by physically interpretable short-window dynamics. Historical replay maps risk scores into alert episodes and quantifies alert duration and false-episode burden. The contribution is an offline-evaluated analytics and validation module designed for integration into a port digital twin.

\end{abstract}
\section{Introduction}
\label{sec:introduction}

Ports need pre-warning information before high water levels disrupt access, equipment use, berth operations, landside movement, and safety-critical work \cite{Asif2025,Nevo2022,Klar2023}. Coastal water levels are shaped by astronomical tide, storm surge, non-tidal residuals, meteorological forcing, riverine influence, and local coastal conditions \cite{Qin2023,Tiggeloven2021}. In a port digital twin, these observations are not useful only as an archive. They are expected to update a cyber-physical state, support monitoring, and feed decision support for operators \cite{Fuller2020,Jones2020,Klar2023}. This paper studies one narrow but operationally important part of that workflow: whether recent local water-level histories can support 12-hour warning-oriented flood pre-warning.

Machine learning (ML) methods have increasingly replaced regression-based approaches for flood forecasting and early warning. Existing work extends across rainfall--runoff prediction, river-level forecasting, flood-warning classification, operational forecasting workflows, and coastal sea-level or storm-surge prediction \cite{Mosavi2018,Asif2025,Munoz2021,Nevo2022,Qin2023,Tiggeloven2021}. Lead-time-dependent calibration studies further show that model behaviour changes with warning horizon \cite{Astagneau2024}. These studies provide the technical foundation, but the task considered here is different from ordinary continuous water-level, river-stage, or surge-residual prediction. The output is not a future water level, but rather a warning-aligned risk score: whether an official flood warning commences within 12 hours from a given local water-level window. This target specification transforms the physical forecasting problem into a sparse, lead-time-specific decision-support problem.

Digital-twin research also changes the form of evaluation. A digital twin is usually described as a data-connected representation that supports monitoring, state update, simulation, and decision support \cite{Fuller2020,Jones2020}. Port digital twins extend this idea to situational awareness and multi-stakeholder coordination \cite{Klar2023}. For flood warning analytics, this means that a model should not be judged only by shuffled row-level classification scores. It should also be judged by whether its risk scores can be converted into alert episodes with usable lead time and acceptable false-alert burden.

The main difficulty is not simply class imbalance. In Liverpool, positive windows constitute 19.0\% of the table, but they arise from only four assigned warning clusters. Most rows describe normal or non-warning conditions, while potentially informative pre-warning windows are concentrated around a few related episodes. Adjacent windows overlap in time and may share the same hydrometeorological context. In data with temporal, spatial, or hierarchical dependence, random cross-validation can underestimate predictive error when related observations are divided between model development and evaluation \cite{Roberts2017}. The same concern applies to feature selection: ranking predictors before the held-out incident is separated can transmit information from that incident into the selected feature set \cite{AmbroiseMcLachlan2002,VarmaSimon2006,Kaufman2012}.

Standard imbalanced-learning methods help with class-frequency skew, but they do not by themselves solve this incident-dependence problem. Oversampling methods such as SMOTE can change the minority-class frequency \cite{Chawla2002}; they do not prevent neighbouring windows from the same warning episode from leaking across train and test folds. Precision--recall evaluation is more suitable than accuracy in imbalanced binary prediction \cite{Saito2015}, but the validation design still has to respect the grouped incident structure.

Sparse feature selection is useful in this setting because it limits the number of predictors that can define a warning rule. Lasso and ElasticNet are established approaches for selection under correlated predictors \cite{Tibshirani1996,ZouHastie2005}, while stability analysis is useful when several descriptors represent similar information \cite{Meinshausen2010}. Stochastic gates and LassoNet provide more flexible nonlinear alternatives \cite{Yamada2020,Lemhadri2021}. The objective here is not to delineate a new sparse-learning algorithm. It is to test whether sparse selection remains useful in a port flood warning context when it is embedded inside incident-cluster validation, as compared against dense non-sparse baselines, and challenged by random feature-subset controls.

We therefore define 12-hour port flood pre-warning as sparse incident-cluster learning. The target is warning-aligned and lead-time-specific; the validation unit is the warning-event cluster; feature selection and threshold selection are nested inside training and validation clusters; and model output is evaluated both as window-level risk scores and as alert episodes. The protocol is intended as a digital-twin analytics module, not as a complete live digital twin.

The empirical study uses Liverpool as the primary case and Humber/Hull proxy and Wessex South as harmonised protocol-transfer checks. In Liverpool, the fold-specific top-10 ElasticNet model reaches an $F_2$ of 0.696, improving on 0.633 for the same model family without top-$k$ truncation and remaining competitive with 0.681 for the full-feature weighted XGBoost reference. The result establishes the top-10 signature as a compact, auditable representation for the alert-oriented objective, while the complete metric profile is reported in the results tables.

This paper makes four contributions. First, it formulates 12-hour port flood pre-warning around the independent warning cluster rather than the individual positive row. Second, it evaluates fold-specific ElasticNet selection without allowing the outer test rows to rank features, and reports the resulting feature stability. Third, it bounds the selected signature through full-feature references, real-versus-synthetic-label controls, 100-repeat random top-$k$ controls, and contextual-versus-physical decomposition. Fourth, it connects window-level scores to digital-twin decision quantities through historical-stream replay, including warning and watch hits, false warning episodes per day, lead time, missed incidents, and alert duration.

\section{Data and Problem Formulation}

\subsection{Data Sources}

Table~\ref{tab:data_sources} summarizes the data sources used in the study. The primary modelling data are derived from Environment Agency water-level and warning records. Additional tide and external-driver sources were explored during the broader study, but the final sparse signature reported here is based on prediction-time available water-window descriptors and contextual covariates.

\begin{table}[!tbp]
\caption{Data sources and their roles in the study. Links are provided through the cited data-source records rather than as raw URLs in the main text.}
\label{tab:data_sources}
\centering
\scriptsize
\begin{tabularx}{\textwidth}{@{}p{0.17\textwidth}p{0.15\textwidth}XXp{0.05\textwidth}@{}}
\toprule
Source & Provider & Variables used & Role in this paper & Ref. \\
\midrule
Flood monitoring archive & Environment Agency & Timestamped observed water levels, monitoring-site descriptors, and measurement-category records & Constructs the eight-point local water-level windows for Liverpool and additional port-region stress tests & \cite{EnvironmentAgencyArchive} \\
Real-time flood monitoring API & Environment Agency & Station metadata, river/sea level station descriptors, and observation metadata & Supports site mapping, measurement-category definition, and reproducible access to level observations & \cite{EnvironmentAgencyAPI} \\
Flood warning information service & Environment Agency & Public flood warning and alert records, warning dates, and warning categories & Defines warning-event dates, incident clusters, and 12-hour pre-warning labels & \cite{EnvironmentAgencyWarnings} \\
Processed tide products and tide-gauge records & BODC and NTSLF & Astronomical tide, observed tide, and derived residual candidates where available & Used in exploratory checks of tide-phase and surge-residual features; not retained as the main reported sparse signature & \cite{BODCNTSLF} \\
\bottomrule
\end{tabularx}
\end{table}

\subsection{Warning-Aligned Datasets}

The primary case study is Liverpool. Additional port-region datasets are used as stress tests of the same protocol rather than as proof of universal geographic generalization. Table~\ref{tab:datasets} summarizes the modelling datasets.

\begin{table}[!tbp]
\caption{Warning-aligned datasets used in the experiments. Clusters are independent warning-event clusters used for grouped validation.}
\label{tab:datasets}
\centering
\scriptsize
\begin{tabularx}{\textwidth}{@{}XrrrrrX@{}}
\toprule
Dataset & Windows & Clusters & Positives & Pos. rate & Stations & Role \\
\midrule
Liverpool primary & 2067 & 4 & 393 & 0.190 & 2 & Main case study \\
Liverpool common-feature view & 2067 & 4 & 393 & 0.190 & 2 & Cross-site baseline \\
Humber/Hull proxy & 4676 & 8 & 766 & 0.164 & 5 & Protocol stress test \\
Wessex South & 2139 & 6 & 327 & 0.153 & 3 & Protocol stress test \\
\bottomrule
\end{tabularx}
\end{table}

The Liverpool table covers aligned source days from September to November 2025. Its four stored warning-cluster labels are C1 (warning dates 7--11 September), C2 (20 September), C3 (6--8 October), and C4 (14 November). These labels are taken from the warning-aligned database and are not re-estimated during model fitting. The 393 positive rows therefore do not constitute 393 independent incidents. Humber/Hull and Wessex South use separately reconstructed warning-aligned tables and a 67-feature common view for the cross-site analysis; their scores are not directly interchangeable with the 71-candidate Liverpool primary analysis.

\subsection{Eight-Point Water-Level Windows}

Let $v_t$ denote the observed water level at the current window endpoint,
after station-level alignment and quality filtering. An inclusive window
containing $L$ observations is defined as
\[
\mathbf{x}_t =
\left(v_{t-(L-1)},v_{t-(L-2)},\ldots,v_t\right).
\]
In this study, $L=8$, and therefore
\[
\mathbf{x}_t =
\left(v_{t-7},v_{t-6},\ldots,v_t\right).
\]
The earliest index is $t-7$, rather than $t-8$, because the current
observation $v_t$ is included together with the previous seven
observations. With a 15-minute sampling interval, the elapsed time
between the first and final observations is
$(L-1)\times15=105$ minutes.

From $\mathbf{x}_t$ we construct prediction-time descriptors that summarize local magnitude, amplitude, short-term increments, directional reversals, local trend, and contrast between the first and second halves of the window. All descriptors are computed using observations available at or before $t$, so no future water-level or warning information is used during feature construction.

\subsection{Target Definition and Incident Clusters}

The prediction target is a 12-hour official-warning pre-warning label. Let $\mathcal{W}=\{T_1,T_2,\ldots,T_J\}$ denote the set of official warning start times, and let $H=12$ hours denote the pre-warning horizon. For a window ending at time $t_i$, the binary label is defined as
\[
y_i =
\begin{cases}
1, & \exists T_j \in \mathcal{W} \text{ such that } t_i < T_j \leq t_i + H,\\
0, & \text{otherwise}.
\end{cases}
\]
Thus, a positive example is not defined by the water level at time $t_i$ itself. It is defined by whether an official flood warning starts within the following 12 hours. This makes the task warning-aligned and lead-time-specific.
When $y_i=1$, the warning event associated with window $i$ is the
earliest warning that starts within the prediction horizon:
\[
j^*(i)=
\arg\min_{j:\,0<T_j-t_i\leq H}
\left(T_j-t_i\right).
\]
Let $\kappa(j)$ denote the stored incident-cluster identifier associated
with warning event $j$. The incident-cluster assignment of a positive
window is then
\[
g_i=\kappa\!\left(j^*(i)\right).
\]
Accordingly, all positive windows associated with the same stored
warning incident share the same group identifier during validation.
Let
\[
\mathcal{C}=\{C_1,C_2,\ldots,C_M\}
\]
denote the set of warning-event clusters. Each positive window is assigned to the stored cluster of the warning event that defines its label. The validation unit is therefore the warning-event cluster rather than the individual row. The present analysis does not infer a new clustering threshold from the outcome data.

In grouped validation, windows whose positive label is assigned to the held-out warning cluster are excluded from model training. Feature ranking is repeated for every outer fold, and the saved selection audit confirms that outer-test rows are not used to rank the top-10 features. The source-day audit found no training--test day overlap, but it did identify one shared source day between validation and test subsets in folds 1--3. Consequently, the design protects the feature-selection boundary and the assigned training--test incident boundary, but it should not be described as complete temporal isolation. Threshold-dependent results are interpreted with this limitation.

\subsection{Feature Taxonomy}

For each window ending at time $t_i$, the model input is written as
\[
\mathbf{z}_i = [\mathbf{c}_i, \mathbf{m}_i, \mathbf{d}_i].
\]
Here, $\mathbf{c}_i$ contains prediction-time contextual covariates: cyclic time-of-day coordinates, monitoring-location identity, and measurement-category identity. The vector $\mathbf{m}_i$ contains the eight observed levels together with level-magnitude summaries such as the local mean, minimum, and maximum. The vector $\mathbf{d}_i$ contains first- and second-order increments, local trend, amplitude, endpoint and half-window contrasts, extremum-position summaries, and directional-reversal descriptors.

For notational clarity, write the eight-point history as
$\mathbf{x}_i=(x_{i,1},\ldots,x_{i,L})$, where $L=8$, and define
\[
\Delta x_{i,k}=x_{i,k+1}-x_{i,k},
\qquad k=1,\ldots,L-1.
\]
The principal physical descriptors are formalised as
\[
\begin{aligned}
\mu_i &=
\frac{1}{L}\sum_{k=1}^{L}x_{i,k},\\
r_i &=
\max_k(x_{i,k})-\min_k(x_{i,k}),\\
d_i^{\max} &=
\max_k|\Delta x_{i,k}|,\\
d_i^{\mathrm{mean}} &=
\frac{1}{L-1}\sum_{k=1}^{L-1}|\Delta x_{i,k}|,\\
q_i &=
\sum_{k=2}^{L-1}
\mathbb{I}\!\left(
\Delta x_{i,k}\Delta x_{i,k-1}<0
\right),\\
h_i &=
\frac{1}{4}\sum_{k=5}^{8}x_{i,k}
-
\frac{1}{4}\sum_{k=1}^{4}x_{i,k}.
\end{aligned}
\]
Here, $q_i$ counts within-window directional reversals and $h_i$
measures the mean-level contrast between the second and first halves
of the window. Using the sample index $k$ for the local trend, let
$\bar{k}=(L+1)/2$. The least-squares slope and second-order increments
are
\[
\begin{aligned}
s_i &=
\frac{\sum_{k=1}^{L}(k-\bar{k})(x_{i,k}-\mu_i)}
{\sum_{k=1}^{L}(k-\bar{k})^2},\\
\Delta^2 x_{i,k}
&= \Delta x_{i,k}-\Delta x_{i,k-1}
= x_{i,k+1}-2x_{i,k}+x_{i,k-1},
\qquad k=2,\ldots,L-1.
\end{aligned}
\]
Thus, $s_i$ measures the signed local trend per sampling step, whereas
$\Delta^2 x_{i,k}$ represents the change in the first-order increment.
\section{Proposed Sparse Pre-warning Analytics Module}

The evaluated module is an analytics protocol rather than a new classifier. Imbalanced-learning methods address class-frequency skew but do not control dependence between overlapping windows from the same warning episode. Sparse selection can produce compact models, but only if feature ranking is separated from the outer test rows. The protocol therefore combines prediction-time feature filtering, fold-specific selection, incident-cluster grouping, random compact-feature controls, and alert-episode evaluation. Its contribution is the evaluation design and decision mapping, not a novel optimisation algorithm.

\subsection{Module Interface in a Port Digital Twin}

\begin{figure}[!tbp] \centering \includegraphics[width=\textwidth,trim=9.8bp 21.0bp 15.7bp 10.0bp,clip]{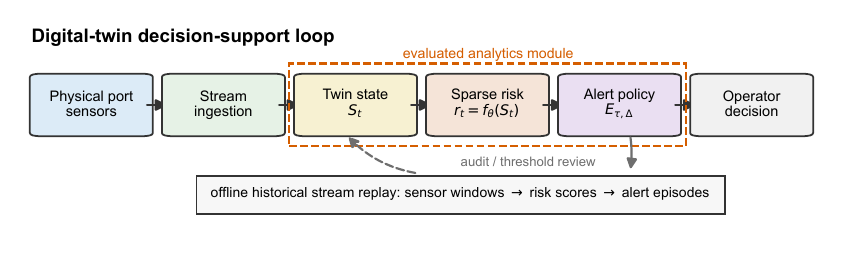} \caption{Digital-twin analytics loop for port flood pre-warning. The evaluated module connects water-level stream ingestion, digital state update, sparse risk inference, alert episode formation, and operator-facing decision support. The study evaluates this loop through offline historical replay rather than live deployment.} \label{fig:digital_twin_loop} \end{figure}

\begin{figure}[!tbp] \centering \includegraphics[width=\textwidth,trim=11.2bp 29.5bp 3.5bp 9.8bp,clip]{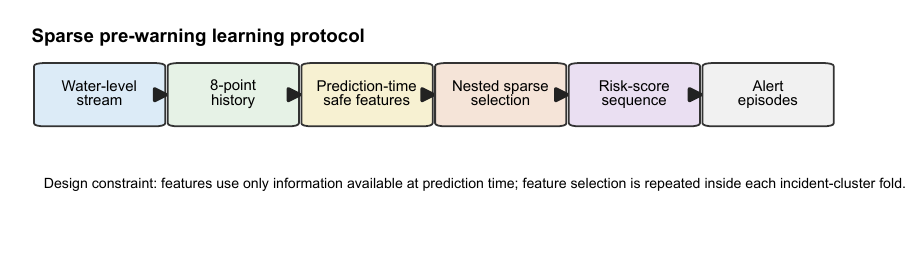} \caption{Sparse pre-warning learning protocol. The protocol converts water-level streams into eight-point histories, prediction-time contextual and physical descriptors, nested sparse feature selection, risk scores, and alert episodes under incident-cluster validation.} \label{fig:sparse_learning_protocol} \end{figure}

Figure~\ref{fig:digital_twin_loop} shows where the proposed analytics module sits inside a port digital-twin workflow. The physical system provides water-level observations, the digital state stores the recent water-level history and context, the analytics module produces a risk score, and the policy layer converts scores into alert episodes. Figure~\ref{fig:sparse_learning_protocol} summarizes the sparse pre-warning learning protocol used to implement this module.

\subsection{Feature Construction}

The feature-construction path is summarized in Fig.~\ref{fig:sparse_learning_protocol}. For each aligned station series, eight-point windows are extracted in chronological order. Prediction-time available features are then computed from the current and previous observations only. Leakage filters remove warning-status fields, future-event information,
and label-derived predictors.

\subsection{Sparse Feature Selection}

The main compact model is elastic-net logistic regression. It estimates the probability that a window belongs to the 12-hour pre-warning class:
\[
P(y_t=1|\mathbf{z}_t)=\sigma(\beta_0+\mathbf{z}_t^\top \boldsymbol{\beta}),
\]
where $\mathbf{z}_t$ is the feature vector and $\sigma(\cdot)$ is the logistic function. For training samples indexed by $i$, the fitted objective is
\[
\min_{\beta_0,\boldsymbol{\beta}}
\sum_i w_i\ell\!\left(y_i,\sigma(\beta_0+\mathbf{z}_i^\top\boldsymbol{\beta})\right)
+\lambda\left(\alpha\|\boldsymbol{\beta}\|_1+
\frac{1-\alpha}{2}\|\boldsymbol{\beta}\|_2^2\right),
\]
where $\ell$ is binary logistic loss and the balanced class weights are represented by $w_i$. The $\ell_1$ component encourages sparsity, while the $\ell_2$ component reduces instability among correlated window descriptors. Categorical variables are imputed by their most frequent training value and one-hot encoded; numerical variables are median-imputed. The saved implementation did not apply an additional numerical scaling stage, so coefficient magnitude is not interpreted as a physical effect size.

Within each fold, $C\in\{0.03,0.1,0.3,1.0\}$ and the elastic-net mixing ratio $\alpha\in\{0.5,0.8,1.0\}$ are selected by validation AP. The final top-10 list is then ranked from the non-zero absolute coefficients using only the model-development subset. Decision thresholds are searched on $\{0.05,0.10,\ldots,0.95\}$ to maximise validation $F_2$, subject to a precision floor of 0.25 when feasible.

The purpose of sparse selection is not to claim that elastic-net is a novel model. Its role is to make the warning signature compact, reproducible, and easier to inspect under sparse incident-cluster validation.

\subsection{Sparse and Non-sparse Comparison Design}

The top-10 model is compared with two full-feature references. The first is an ElasticNet logistic model fitted to all 71 Liverpool candidate variables without top-$k$ truncation; this comparison holds the classifier family fixed. The second, denoted Current12hBinary, is a full-feature weighted XGBoost classifier with a binary 12-hour target. It is an established nonlinear reference rather than a controlled sparsity ablation. A top-10 XGBoost model tests whether nonlinear interactions improve the same compact representation.

All models use the same 12-hour target, outer warning-cluster assignments, validation-selected threshold rule, and evaluation metrics. The full- versus top-10 ElasticNet comparison isolates top-$k$ truncation within one model family. Comparisons with Current12hBinary additionally reflect differences between logistic and boosted-tree classifiers and are interpreted as reference comparisons rather than causal evidence for sparsity.

\subsection{Negative and Random-Subset Controls}

Two types of controls constrain interpretation. First, real warning labels are compared with day-group permutations, temporally shifted warnings, and pseudo-warning days sampled from normal periods. These controls test whether performance follows official warning timing rather than arbitrary event placement. Second, eight random feature-subset schemes are repeated 100 times. Each repeat runs the complete four-fold grouped protocol, so the repeat-level metric is the mean across the four outer folds. We also retain the distribution of individual fold results to expose event-specific high tails. Random controls include all-feature, no-time, no-station, physical-only, context-only, matched-family, and matched-alert-budget subsets.

\subsection{Nested Grouped Validation}

\begin{figure}[!tbp]
\centering
\includegraphics[width=\textwidth,trim=11.6bp 29.5bp 3.6bp 9.8bp,clip]{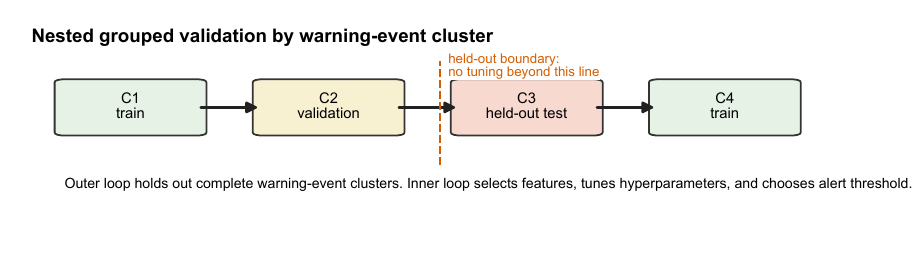}
\caption{Incident-cluster validation design. Outer folds hold out assigned warning-event clusters, and feature ranking does not use outer-test rows. A source-day audit found no training--test overlap but found one validation--test source-day overlap in folds 1--3; threshold-dependent metrics are therefore reported with this caveat.}
\label{fig:validation}
\end{figure}

Figure~\ref{fig:validation} illustrates the validation design. The outer loop holds out one assigned warning cluster. A second cluster supplies model selection and threshold tuning, and the remaining clusters supply training rows. Feature ranking records confirm that outer-test rows were not used. Because the stored source-day audit detected validation--test overlap in three folds, AP is the least threshold-dependent primary metric and all threshold-derived scores are treated as descriptive estimates rather than leakage-free generalisation bounds.

\subsection{Metrics}

Average precision (AP) evaluates ranking quality without choosing a decision threshold and is more informative than accuracy for an imbalanced warning target. We also report the standard $F_2$ score at the validation-selected threshold:
\[
F_2 = \frac{5PR}{4P+R},
\]
where $P$ is precision and $R$ is recall. The $F_2$ score weights recall more heavily than precision, which reflects the operational cost of missed warning opportunities. The term ``validation-threshold $F_2$'' is used below; it is not a new mathematical metric.

Event recall is computed from distinct warning timestamps present in each held-out fold and then averaged across folds; it is not the same denominator as the four assigned incident clusters. Row-level false alarms per day count false-positive windows divided by represented source days. Episode-level metrics instead merge sequential positive windows and report warning hit rate, watch hit rate, median lead time in hours, missed warning instances, and mean alert duration in hours.

\subsection{Baselines and Controls}

The baseline set separates three questions. Dense full-feature baselines test whether ordinary non-sparse learning already solves the task when all safe predictors are available. Context-only and physical-only models test whether the signal comes mainly from temporal/site context or from water-window dynamics. Random top-$k$ controls test whether the selected top-10 signature is more informative than arbitrary compact subsets of the same size.

The key random controls are random all-feature top-10, random no-time top-10, random physical-only top-10, random context-only top-10, and random subsets with the same family composition as the nested selected signature. Negative controls test whether real warning labels outperform deliberately weakened label or feature constructions.

\subsection{Alert Episode Policy}

Window-level risk scores are converted into alert episodes using score aggregation and threshold policies. Consecutive alerting windows are merged into an episode. We evaluate three hit rules:
\[
\text{watch hit}: [T_w-24h,T_w-12h],
\]
\[
\text{warning hit}: [T_w-12h,T_w],
\]
and
\[
\text{any pre-event hit}: [T_w-24h,T_w).
\]
This separates model ranking from operational burden. A model with high recall but very long alert episodes may be less useful for decision support.

\section{Offline Twin Replay Evaluation}

To connect model evaluation with digital-twin decision support, we replay each held-out fold in timestamp order. This is an experimental protocol, not a live deployment, a hydrodynamic simulation, or a strict forward-chaining evaluation. Model development follows the grouped folds described above and may include incidents later than a held-out cluster; only the held-out score sequence itself is replayed chronologically. The purpose is to evaluate the decision objects that the analytics module would emit from sequential observations.

\begin{figure}[!tbp] \centering \includegraphics[width=\textwidth,trim=11.4bp 32.8bp 3.6bp 9.8bp,clip]{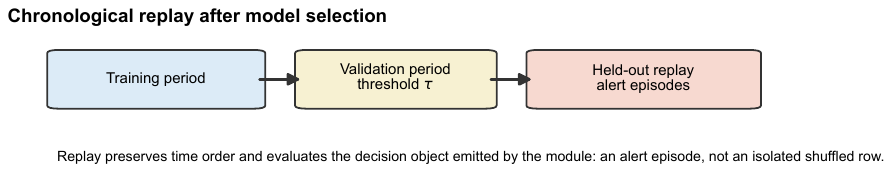} \caption{Foldwise historical replay design. Held-out water-level windows are scored in timestamp order and converted into alert episodes. The replay preserves the order of the evaluated stream but is distinct from a strict latest-event temporal holdout.} \label{fig:chronological_replay} \end{figure}

Formally, let $\{(\mathbf{x}_t,\mathbf{c}_t)\}_{t=1}^{T}$ denote the ordered sequence of water-level histories and contextual covariates in a held-out fold. A trained model $f_{\theta}$ produces $s_t=f_{\theta}(\mathbf{x}_t,\mathbf{c}_t)$. Scores are aggregated by their maximum across simultaneous station--measurement windows. The threshold $\tau$ is the stored validation-selected $F_2$ threshold. An episode operator $\mathcal{E}_{\tau,\Delta}$ thresholds the sequence and merges alerts separated by no more than $\Delta=30$ minutes; one positive window is sufficient to start an episode. The resulting episodes are matched to official warning timestamps only after the full held-out sequence has been generated. Thresholds obtained by scanning the evaluation fold for a false-episode budget are excluded from the reported primary policy because they are not deployment-valid.
Figure~\ref{fig:chronological_replay} illustrates the foldwise replay design. Unlike shuffled row-level evaluation, the replay keeps the historical ordering of water-level windows within each held-out fold, updates the digital-twin state using only information available up to the current replay time, emits risk scores sequentially, and then converts these scores into alert episodes before matching them to official warning times.
\begin{table}[!tbp]
\caption{Offline twin replay protocol for evaluating alert episodes from held-out historical streams.}
\label{tab:offline_replay_protocol}
\centering
\footnotesize
\begin{tabularx}{\textwidth}{p{0.12\textwidth}X}
\toprule
Stage & Operation \\
\midrule
Input & Timestamp-ordered held-out water-level histories, prediction-time contextual covariates, trained risk model $f_{\theta}$, validation-selected threshold $\tau$, and a 30-minute episode-merging gap. \\
State update & At replay time $t$, update the digital state using only observations available at or before $t$. \\
Risk inference & Compute a pre-warning risk score $s_t=f_{\theta}(\mathbf{x}_t,\mathbf{c}_t)$ for the current replay state. \\
Episode formation & Aggregate simultaneous scores by their maximum, apply $\mathcal{E}_{\tau,\Delta}$, and merge alerts separated by at most 30 minutes. \\
Event matching & Compare completed alert episodes with official warning times using watch-window, warning-window, and any-pre-event hit rules. \\
Output & Report warning hit rate, watch hit rate, false warning episodes per day, median lead time, missed incidents, and mean alert duration. \\
\bottomrule
\end{tabularx}
\end{table}

This replay is closer to a digital-twin decision interface than shuffled row evaluation because it separates risk inference from episode formation and evaluates an operator-facing alert object. It does not establish prospective temporal generalisation or operational readiness.

\section{Results}

\subsection{Main Predictive Performance and Cluster Sensitivity}
Table~\ref{tab:main_results} separates a controlled top-$k$ comparison from the nonlinear reference comparison. The all-candidate and top-10 ElasticNet rows use the same classifier family, whereas Current12hBinary is a weighted XGBoost model. Means and standard deviations are calculated across the four Liverpool folds.

\begin{table}[!tbp]
\caption{Liverpool comparison under grouped incident-cluster evaluation. Values are mean $\pm$ standard deviation across four folds. ``All candidates'' denotes input variables before ElasticNet shrinkage; ``top-10'' denotes fold-specific truncation after ElasticNet ranking; ``weighted XGBoost (full)'' corresponds to Current12hBinary.}
\label{tab:main_results}
\centering
\footnotesize
\setlength{\tabcolsep}{3.5pt}
\begin{tabularx}{\textwidth}{Xrrrrr}
\toprule
Model & AP & \sparseftwo & Event recall & False/day & Inputs \\
\midrule
Weighted XGBoost (full) & $0.577\pm0.202$ & $0.681\pm0.125$ & 1.000 & 33.41 & 71 \\
ElasticNet (all candidates) & $0.592\pm0.160$ & $0.633\pm0.160$ & 0.708 & 17.76 & 71 \\
ElasticNet (fold-specific top-10) & $0.567\pm0.169$ & $\mathbf{0.696}\pm0.089$ & 0.833 & 24.72 & 10 \\
XGBoost (fold-specific top-10) & $0.553\pm0.171$ & $0.650\pm0.136$ & 0.917 & 37.56 & 10 \\
ElasticNet (context only) & $0.417\pm0.183$ & $0.679\pm0.119$ & 0.729 & 25.73 & 4 \\
ElasticNet (physical only) & $0.383\pm0.056$ & $0.587\pm0.160$ & 0.813 & 28.54 & 36 \\
\bottomrule
\end{tabularx}
\end{table}

\begin{figure}[!tbp]
\centering
\includegraphics[width=0.88\textwidth]{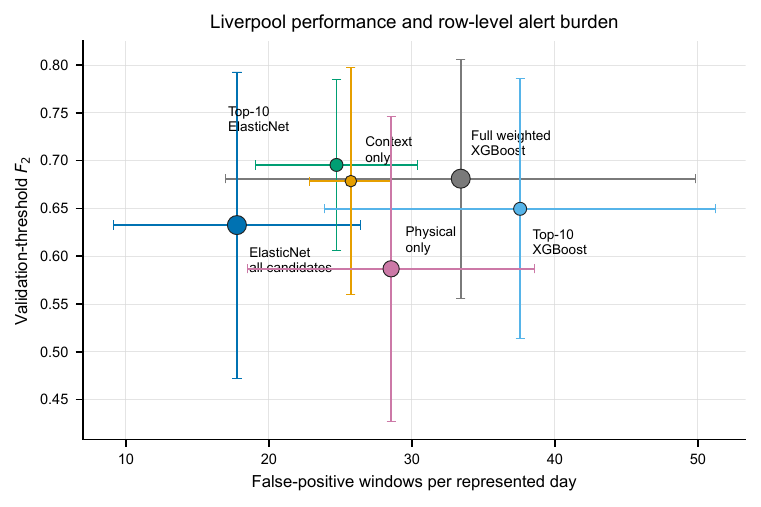}
\caption{Liverpool mean validation-threshold $F_2$ and row-level false-alarm burden. Horizontal and vertical error bars are fold standard deviations. Marker area encodes the number of candidate inputs.}
\label{fig:performance}
\end{figure}

Top-10 truncation raises mean validation-threshold $F_2$ from 0.633 to 0.696 within the ElasticNet model family while reducing the candidate representation from 71 variables to 10 selected predictors. Against the weighted XGBoost reference, the compact model achieves a comparable $F_2$ and fewer false-positive windows per represented day. Table~\ref{tab:main_results} retains AP, event recall, and false-positive burden so that the $F_2$-oriented gain is interpreted within the complete metric profile.

Table~\ref{tab:fold_sensitivity} shows why the four-cluster mean is insufficient on its own. The top-10 model is strongest on C1, while C2 has the lowest AP and detects only half of the warning timestamps represented in that fold. Thresholds also vary from 0.05 to 0.55, indicating sensitivity to the small validation sets.

\begin{table}[!tbp]
\caption{Fold-level Liverpool results for the top-10 ElasticNet model, with the full weighted XGBoost $F_2$ shown as a reference.}
\label{tab:fold_sensitivity}
\centering
\footnotesize
\begin{tabular}{llrrrrrr}
\toprule
Fold & Test cluster & Threshold & AP & \sparseftwo & Event recall & False/day & XGB $F_2$ \\
\midrule
1 & C1: 7--11 Sep. & 0.35 & 0.796 & 0.826 & 1.000 & 20.57 & 0.864 \\
2 & C2: 20 Sep. & 0.55 & 0.441 & 0.624 & 0.500 & 19.17 & 0.631 \\
3 & C3: 6--8 Oct. & 0.05 & 0.592 & 0.668 & 0.833 & 28.82 & 0.649 \\
4 & C4: 14 Nov. & 0.25 & 0.438 & 0.664 & 1.000 & 30.33 & 0.581 \\
\bottomrule
\end{tabular}
\end{table}

\subsection{Selection and Label Controls}
Table~\ref{tab:random_controls} reports 100 complete repetitions per control, where every repetition aggregates the same four-fold evaluation used for the selected model. This repeat-level summary should be distinguished from the distribution of 400 individual fold results.

\begin{table}[!tbp]
\caption{Repeat-level random-subset controls ($n=100$ complete four-fold repetitions per row). The selected top-10 reference has AP 0.567 and validation-threshold $F_2$ 0.696.}
\label{tab:random_controls}
\centering
\footnotesize
\setlength{\tabcolsep}{3.7pt}
\begin{tabularx}{\textwidth}{Xrrrr}
\toprule
Control & AP mean $\pm$ SD & $F_2$ mean $\pm$ SD & $F_2$ p95 & Runs $\geq$ selected \\
\midrule
All candidates & $0.360\pm0.033$ & $0.553\pm0.049$ & 0.624 & 0\% \\
No cyclic time & $0.336\pm0.017$ & $0.531\pm0.043$ & 0.586 & 0\% \\
No station identity & $0.364\pm0.037$ & $0.550\pm0.055$ & 0.631 & 0\% \\
No time or station & $0.335\pm0.017$ & $0.543\pm0.041$ & 0.596 & 0\% \\
Physical candidates only & $0.347\pm0.012$ & $0.528\pm0.032$ & 0.580 & 0\% \\
Context candidates only ($k=4$) & $0.417\pm0.000$ & $0.685\pm0.000$ & 0.685 & 0\% \\
Matched feature-family composition & $0.476\pm0.027$ & $0.678\pm0.009$ & 0.691 & 1\% \\
Matched feature count and alert budget & $0.362\pm0.037$ & $0.379\pm0.101$ & 0.518 & 0\% \\
\bottomrule
\end{tabularx}
\end{table}

\begin{figure}[!tbp]
\centering
\includegraphics[width=0.88\textwidth]{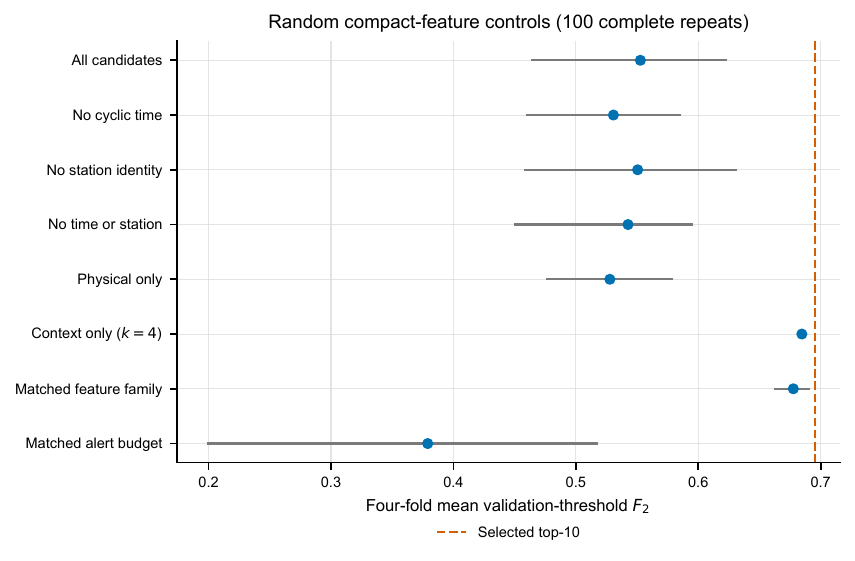}
\caption{Repeat-level random top-10 controls. Points show mean validation-threshold $F_2$, horizontal bars show the 5th--95th percentile range, and the dashed line is the selected top-10 result (0.696).}
\label{fig:random_controls}
\end{figure}

The selected result exceeds the repeat-level 95th percentile of every control, although 1\% of matched-family repetitions equal or exceed it. At the individual-fold level, the all-candidate and matched-family 95th percentiles are 0.748 and 0.847, respectively. This difference shows that a subset can perform unusually well on one incident without remaining strong after all four folds are aggregated. The selected predictors are therefore more reliable than broad random subsets at the protocol level, but they are not a unique physical law.

Real warning timing also outperforms synthetic controls. Table~\ref{tab:negative_controls} compares real-label AP with the largest AP obtained by day-group permutation or pseudo-warning controls. The 24-hour and 48-hour warning shifts provide additional checks but are not used to tune the real-label model.

\begin{table}[!tbp]
\caption{Real warning labels versus the strongest synthetic negative control for each model.}
\label{tab:negative_controls}
\centering
\footnotesize
\begin{tabular}{lrrr}
\toprule
Model & Real-label AP & Maximum control AP & AP difference \\
\midrule
Full weighted XGBoost & 0.580 & 0.406 & 0.173 \\
Top-10 ElasticNet & 0.567 & 0.367 & 0.200 \\
Top-10 XGBoost & 0.553 & 0.379 & 0.174 \\
\bottomrule
\end{tabular}
\end{table}

\subsection{Composition and Stability of the Sparse Signature}
Table~\ref{tab:context_physics} separates cyclic time, station identity, physical level, physical shape, the fold-specific hybrid top-10, and the all-candidate ElasticNet model. All rows use the same classifier family and fold definitions.

\begin{table}[!tbp]
\caption{Context-versus-physics decomposition using ElasticNet logistic regression. Values are four-fold means.}
\label{tab:context_physics}
\centering
\footnotesize
\begin{tabular}{lrrrr}
\toprule
Signature & Inputs & \sparseftwo & AP & Event recall \\
\midrule
Cyclic time only & 2 & 0.678 & 0.418 & 0.792 \\
Station identity only & 1 & 0.618 & 0.253 & 1.000 \\
Physical level only & 11 & 0.554 & 0.350 & 1.000 \\
Physical shape only & 25 & 0.594 & 0.380 & 0.813 \\
All physical, no context & 36 & 0.587 & 0.383 & 0.813 \\
Fold-specific hybrid top-10 & 10 & \textbf{0.696} & 0.567 & 0.833 \\
All candidates, no top-$k$ truncation & 71 & 0.633 & \textbf{0.592} & 0.708 \\
\bottomrule
\end{tabular}
\end{table}

\begin{figure}[!tbp]
\centering
\includegraphics[width=0.88\textwidth]{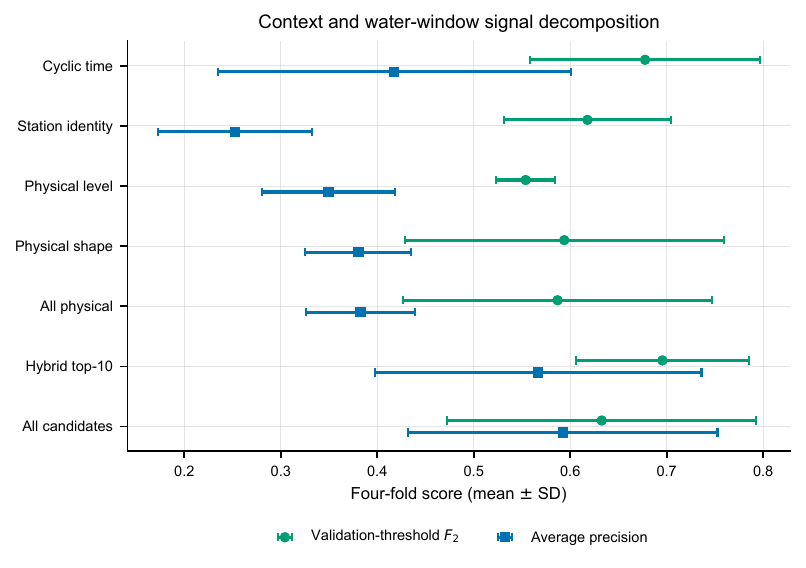}
\caption{Context-versus-physics decomposition. Points show four-fold means and error bars show fold standard deviations. AP and validation-threshold $F_2$ are displayed as distinct metrics and should not be compared as if they shared the same decision threshold.}
\label{fig:context_physics}
\end{figure}

The cyclic-time model provides a strong contextual baseline, while physical descriptors retain predictive value without contextual covariates. The selected Liverpool signature combines both sources: temporal and site context anchor the warning state, and short-window level dynamics contribute physically interpretable variation. This decomposition supports a hybrid, context-anchored representation; statistical separation of the physical component after context adjustment remains a future inferential question.

The selected signature is fold-specific rather than one universal list. Table~\ref{tab:feature_stability} reports predictors appearing in at least two of the four top-10 lists. Cyclic time, site identity, directional reversal count, and maximum absolute increment appear in every fold; several endpoint and positional descriptors recur less consistently.

\begin{table}[!tbp]
\caption{Recurring predictors in the fold-specific top-10 ElasticNet lists. Names describe the scientific quantity rather than the implementation column.}
\label{tab:feature_stability}
\centering
\footnotesize
\begin{tabularx}{\textwidth}{Xlr}
\toprule
Predictor & Family & Selected folds \\
\midrule
Cyclic time-of-day cosine & Temporal context & 4/4 \\
Cyclic time-of-day sine & Temporal context & 4/4 \\
Monitoring-location identity & Site context & 4/4 \\
Within-window directional reversal count & Physical shape & 4/4 \\
Maximum absolute 15-minute level increment & Physical shape & 4/4 \\
Net change over the local window & Physical shape & 3/4 \\
Endpoint trend over the local window & Physical shape & 3/4 \\
Position of the local maximum & Physical shape & 2/4 \\
Position of the local minimum & Physical shape & 2/4 \\
Minimum level within the window & Physical level & 2/4 \\
Terminal water-level observation & Physical level & 2/4 \\
Start-to-end level contrast & Physical shape & 2/4 \\
\bottomrule
\end{tabularx}
\end{table}

Table~\ref{tab:physical_sanity} reports physically interpretable water-window predictors. Pre-warning windows show larger local changes and ranges than normal windows.

\begin{table}[!tbp]
\caption{Physical sanity-check predictors comparing normal and pre-warning windows. Standardized mean difference and Cliff's delta summarize whether interpretable water-window variables separate the two states.}
\label{tab:physical_sanity}
\centering
\scriptsize
\begin{tabularx}{\textwidth}{@{}>{\raggedright\arraybackslash}Xrrrrr@{}}
\toprule
Feature & \shortstack{Normal\\mean} & \shortstack{Pre-warning\\mean} & SMD & \shortstack{Cliff's\\delta} & \shortstack{Direction\\consistency} \\
\midrule
Maximum absolute level increment & 0.3134 & 0.4236 & 0.8290 & 0.4215 & 0.75 \\
Local amplitude range & 1.6028 & 2.1678 & 0.5976 & 0.3117 & 0.75 \\
Mean absolute level increment & 0.2332 & 0.3151 & 0.6319 & 0.3316 & 0.75 \\
\bottomrule
\end{tabularx}
\end{table}

These marginal checks do not account for overlap among adjacent windows and do not establish causality or conditional independence from time and station context. They support a narrower claim: local water-level dynamics are physically plausible components of the warning signature.

\subsection{Cross-Site Protocol Stress Test}

Table~\ref{tab:cross_site} reports a separate harmonised experiment using 67 features available at all three sites. It must not be compared numerically with the 71-candidate Liverpool primary experiment. The purpose is to test whether the feature-construction and grouped-evaluation protocol can be repeated, not whether Liverpool model parameters transfer unchanged.

\begin{table}[!tbp]
\caption{Cross-site protocol stress test using the common 67-feature view. Values are mean validation-threshold $F_2$ across site-specific warning-cluster folds.}
\label{tab:cross_site}
\centering
\footnotesize
\begin{tabularx}{\textwidth}{Xrrrrr}
\toprule
Site & \shortstack{Full\\XGBoost} & \shortstack{Port-nested\\top-10} & \shortstack{Liverpool-fixed\\top-10} & \shortstack{Context\\only} & \shortstack{Physical\\sanity} \\
\midrule
Liverpool & 0.381 & 0.481 & 0.457 & 0.558 & 0.509 \\
Humber/Hull proxy & 0.522 & 0.526 & 0.527 & 0.542 & 0.465 \\
Wessex South & 0.503 & 0.527 & 0.514 & 0.528 & 0.422 \\
\bottomrule
\end{tabularx}
\end{table}

Context-only models establish a strong operational baseline at all three sites, and the port-specific top-10 model remains close to the full common-feature XGBoost reference in Humber/Hull and Wessex South. Variation across the locally selected signatures motivates site-specific calibration within a shared feature-construction and incident-cluster validation protocol.

\subsection{Offline Replay and Alert Burden}

Table~\ref{tab:episode_policy} reports the maximum-score aggregation with the threshold stored from validation $F_2$. The table excludes policies whose threshold was selected by scanning the evaluation fold against a false-episode budget. Nineteen warning-timestamp instances are accumulated across the four held-out folds; this denominator differs from the four assigned warning clusters.
\begin{table}[!tbp]
\caption{Foldwise historical replay with maximum-score aggregation and stored validation thresholds. Lead time and duration are reported in hours.}
\label{tab:episode_policy}
\centering
\scriptsize
\setlength{\tabcolsep}{3.2pt}
\renewcommand{\arraystretch}{0.90}

\begin{tabularx}{\textwidth}{@{}>{\raggedright\arraybackslash}Xrrrrrr@{}}
\toprule
Model setting 
& \shortstack{Warning\\hit}
& \shortstack{Watch\\hit}
& \shortstack{False warn.\\/day}
& \shortstack{Median\\lead (h)}
& \shortstack{Mean\\duration (h)}
& Missed \\
\midrule

\shortstack[l]{Context-only\\SparseLogistic}
& 1.000 & 1.000 & 0.564 & 15.28 & 14.75 & 0 \\

\shortstack[l]{Full 12-h\\XGBoost}
& 1.000 & 0.813 & 0.524 & 15.99 & 28.29 & 0 \\

\shortstack[l]{Nested top-10\\SparseLogistic}
& 1.000 & 1.000 & 0.658 & 14.73 & 13.49 & 0 \\

\shortstack[l]{Nested top-10\\XGBoost}
& 1.000 & 0.813 & 0.465 & 16.29 & 29.92 & 0 \\

\shortstack[l]{Physical-only\\SparseLogistic}
& 0.813 & 0.250 & 0.738 & 15.85 & 18.01 & 3 \\

\bottomrule
\end{tabularx}

\vspace{1mm}
\begin{minipage}{\textwidth}
\raggedright\footnotesize
\emph{Note:} SparseLogistic denotes elastic-net sparse logistic regression.
Top-10 denotes the nested top-10 selected feature set.
\end{minipage}
\end{table}
\begin{figure}[!tbp]
\centering
\includegraphics[width=0.88\textwidth]{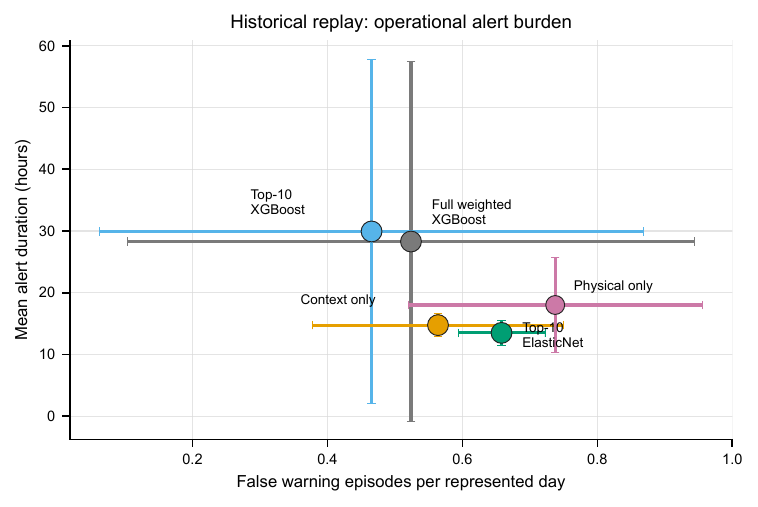}
\caption{Operational burden under stored validation thresholds. Marker position compares false warning episodes per day with mean alert duration; marker area represents warning hit rate.}
\label{fig:episode_policy}
\end{figure}

The full weighted XGBoost, top-10 ElasticNet, top-10 XGBoost, and context-only policies hit every represented warning instance. Their operating points span 13.5--29.9 hours in mean alert duration and 0.47--0.66 false warning episodes per represented day. Historical replay therefore supplies the additional burden coordinates needed to choose an operational threshold rather than relying on window-level AP or $F_2$ alone.

\FloatBarrier

\section{Conclusion and Discussion}

This study evaluates a compact warning-analytics protocol for the case in which many overlapping windows arise from few independent warning clusters. The Liverpool result is not a single-model performance claim. It combines fold-specific feature selection, full-feature references, fold sensitivity, negative-label controls, random feature subsets, contextual--physical decomposition, cross-site protocol checks, and alert-episode replay.

The module updates a short water-level state, estimates 12-hour warning risk, and converts the score sequence into alert episodes. These functions correspond to an analytics component within a port digital twin. Live ingestion, synchronization latency, sensor-failure handling, operator interaction, and prospective deployment are outside the present evaluation.

Sparsity acts as a constraint on the warning rule. Within the ElasticNet family, top-10 truncation raises mean validation-threshold $F_2$ from 0.633 to 0.696 and produces a signature that can be audited across folds. Relative to the weighted XGBoost reference, the compact model remains competitive in $F_2$ while using a substantially smaller representation and generating fewer row-level false-positive windows. The contribution is therefore a metric-specific gain in alert-oriented discrimination together with compactness and auditability.

The random controls refine this interpretation. At the complete-repeat level, the selected model exceeds the 95th percentile of all tested random schemes; at the individual-fold level, context-rich random subsets have broad high tails. The selected list is therefore reproducible as a protocol-level compact representation, but it should not be interpreted as the only correct feature set or as a discovered hydrodynamic law.

\subsection{Context, Tide, and Physical Window Dynamics}

The strong cyclic-time and station results are central to interpretation. In a coastal port, these variables may encode astronomical tidal timing, site exposure, observation practice, season, or warning-reporting regularity. Because verified astronomical phase and surge residual are not retained in the primary signature, the contextual component cannot be assigned uniquely to tidal physics.

Physical short-window descriptors provide consistent supporting evidence. Larger local amplitude, maximum increment, and mean increment are observed in pre-warning windows, and maximum increment plus directional reversals recur in the top-10 lists. These predictive associations establish physical plausibility for the selected signature; a future context-adjusted analysis can quantify their conditional contribution more precisely.

\subsection{Validation Scope and Deployment Path}

The evidence is scoped by four design conditions. Liverpool provides four assigned warning clusters, with fold-level AP ranging from 0.438 to 0.796. The stored split audit reports no training--test source-day overlap and one validation--test source-day overlap in folds 1--3; feature ranking remains separated from outer-test rows, while threshold-dependent metrics are interpreted as protocol estimates. Foldwise replay preserves evaluation order and supplies an intermediate step toward a latest-event prospective test. Finally, the observed 13.5--29.9-hour alert durations define a concrete operating range for future threshold and persistence calibration. These conditions specify the next stages required for deployment-oriented validation.

The cross-site common-feature experiment shows that the workflow can be reconstructed at additional regions and that site-specific calibration is valuable when contextual baselines are strong. Humber/Hull is treated as a proxy-region protocol test rather than evidence of universal model transfer.

\subsection{Summary of Evidence}

This paper presents a sparse pre-warning analytics module for 12-hour port flood warning support under rare warning events. The module represents recent observations as eight-point water-level windows, retains only prediction-time available predictors, selects compact sparse signatures under grouped incident-cluster validation, and evaluates decisions through offline alert episode replay.

The evidence supports four conclusions. First, fold-specific top-10 ElasticNet selection provides the strongest mean $F_2$ among the ElasticNet variants and remains competitive with the full weighted XGBoost reference using ten selected predictors. Second, negative-label and complete-repeat random controls associate the result with real warning timing and place it above broad random compact subsets. Third, the selected representation combines a strong temporal/site context anchor with physically plausible water-window dynamics. Fourth, foldwise stream replay translates scores into digital-twin decision objects and quantifies warning hits together with false episodes and time under alert.

The contribution is therefore a transparent analytics and evaluation protocol for sparse warning-event settings. Larger independent incident archives, verified tide and surge-residual inputs, strict forward temporal validation, and operator-tested alert policies are required before operational deployment.

\begin{credits}
\subsubsection{\ackname}
This work was supported by the UKRI EPSRC project "CRDT-Port: Developing a Crisis and Resilience Digital Twin for UK Port-centric Transport Systems" under Grant UKRI858. This study uses Environment Agency flood and river level data and public flood warning information. The analysis was conducted as an offline historical replay and does not represent an operational warning service.
\subsubsection{\discintname}
The authors declare no competing interests.
\end{credits}

\bibliographystyle{unsrt}
\bibliography{ref}

\end{document}